\documentclass[10pt,twocolumn,letterpaper]{article}

\usepackage[pagenumbers]{wacv} 

\usepackage{bm}
\usepackage{graphicx}
\usepackage{amsmath}
\usepackage{amssymb}
\usepackage{booktabs}
\usepackage{caption}
\usepackage{stfloats}

\usepackage{multirow}
\usepackage{color}
\usepackage{xcolor}
\usepackage[table]{colortbl}
\usepackage{threeparttable}

\usepackage[pagebackref,breaklinks,colorlinks]{hyperref}

\usepackage[capitalize]{cleveref}
\crefname{section}{Sec.}{Secs.}
\Crefname{section}{Section}{Sections}
\Crefname{table}{Table}{Tables}
\crefname{table}{Tab.}{Tabs.}

\def\wacvPaperID{411} 
\def\confName{WACV}
\def\confYear{2026}

\usepackage{soul}
\usepackage{hyperref}
\usepackage{xcolor} 
\newcommand{\hlred}[1]{\sethlcolor{red!35}\hl{#1}}
\newcommand{\hlyellow}[1]{\sethlcolor{yellow!35}\hl{#1}}
\newcommand{\hlorange}[1]{\sethlcolor{orange!35}\hl{#1}}

\newcommand{\zthuang}[1]{}
\newcommand{\zbhe}[1]{}
\newcommand{\dwu}[1]{}
\newcommand{\ml}[1]{}
\newcommand{\mlc}[1]{}
\newcommand{\zt}[1]{}

\begin{document}

\title{Gaussian Set Surface Reconstruction through Per-Gaussian Optimization}

\author{Zhentao Huang\\
{\small School of Computer Science, University of Guelph}\\
{\tt\small zhentao@uoguelph.ca}
\and
Di Wu\\
{\small Faculty of Science and Technology, University of Macau}\\
{\tt\small diwu96063@gmail.com}
\and
Zhenbang He\\
{\small Irving K. Barber Faculty of Science, UBC Okanagan}\\
{\tt\small zbhe96@student.ubc.ca}
\and
Minglun Gong\\
{\small School of Computer Science, University of Guelph}\\
{\tt\small minglun@uoguelph.ca}
}

\twocolumn[{
    \renewcommand\twocolumn[1][]{#1}%
    \maketitle

    \begin{center}
    \includegraphics[width=0.9\textwidth]{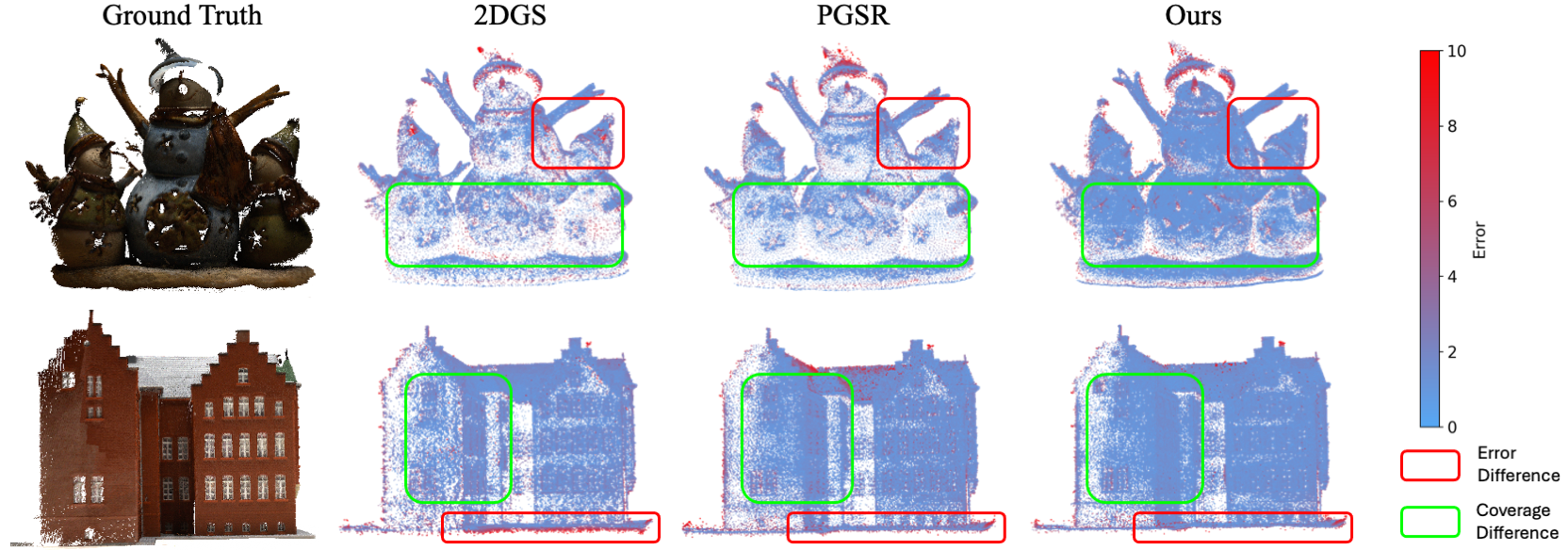}  
    \captionof{figure}{Comparison of Gaussian primitives distributions from 2DGS, PGSR, and our method across two scenes. Gaussian centers are visualized as colored points, with hue indicating surface distance error. Areas where the white background is visible through the Gaussians denote incomplete surface coverage. Red and green boxes highlights regions with improved accuracy and coverage, respectively. Quantitative and visual results confirm that our approach achieves both a lower reconstruction error and a more complete Gaussian representation.}
    \label{fig:teaser}
    \end{center}
}]

\maketitle


\begin{abstract}
    3D Gaussian Splatting (3DGS) effectively synthesizes novel views through its flexible representation, yet fails to accurately reconstruct scene geometry. While modern variants like PGSR introduce additional losses to ensure proper depth and normal maps through Gaussian fusion, they still neglect individual placement optimization. This results in unevenly distributed Gaussians that deviate from the latent surface, complicating both reconstruction refinement and scene editing. 
    Motivated by pioneering work on Point Set Surfaces, we propose Gaussian Set Surface Reconstruction (GSSR), a method designed to distribute Gaussians evenly along the latent surface while aligning their dominant normals with the surface normal. GSSR enforces fine-grained geometric alignment through a combination of pixel-level and Gaussian-level single-view normal consistency and multi-view photometric consistency, optimizing both local and global perspectives. To further refine the representation, we introduce an opacity regularization loss to eliminate redundant Gaussians and apply periodic depth- and normal-guided Gaussian reinitialization for a cleaner, more uniform spatial distribution.
    Our reconstruction results demonstrate significantly improved geometric precision in Gaussian placement, enabling intuitive scene editing and efficient generation of novel Gaussian-based 3D environments. Extensive experiments validate GSSR’s effectiveness, showing enhanced geometric accuracy while preserving high-quality rendering performance. \textbf{Our code will be released upon acceptance of the paper.}

\end{abstract}
\begin{figure*}
    \centering
    \includegraphics[width=.9\textwidth]{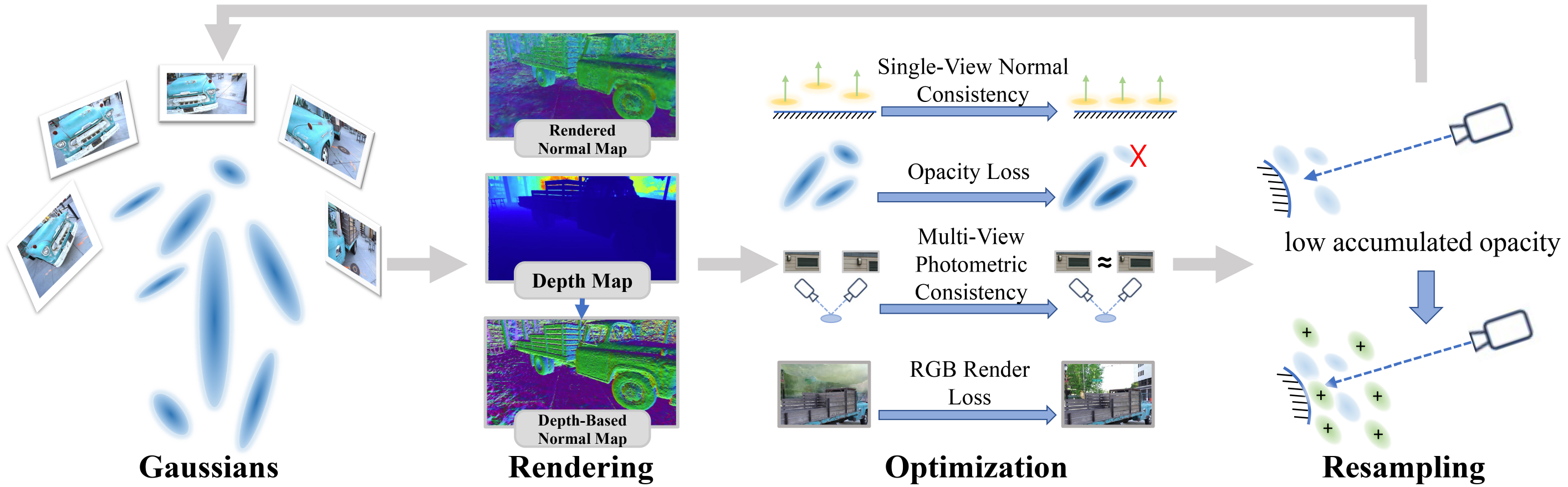}
    \caption{Overview of the proposed GSSR pipeline. Given multi-view posed images and initialized 3D Gaussians, we render the depth map, depth-based normals and alpha-blended normals. The optimization stage includes four major components: (1) single-view normal consistency, (2) multi-view photometric consistency (3) RGB rendering loss, and (4) opacity regularization. Additionally, Gaussians are periodically resampled using our view-based opacity-guided strategy, resulting in a more uniform and accurate distribution.}
    \label{fig:enter-label}
\end{figure*}
\section{Introduction}
\label{sec:intro}

Novel View Synthesis (NVS) remains a fundamental challenge in computer vision and graphics, enabling photorealistic scene rendering from sparse inputs for applications ranging from eXtended Reality (XR) to autonomous systems. Modern NVS approaches can be broadly categorized by their treatment of scene geometry: explicit modeling methods that reconstruct accurate 3D surfaces \cite{schonberger2016sfm,schonberger2016pixelwise_mvs} and implicit scene representations that bypass explicit reconstruction \cite{lightfield96,mildenhall2021nerf, kerbl20233d}.

With advances in deep neural networks, research on implicit scene representations has gained momentum. In particular, Neural Radiance Fields (NeRF) \cite{mildenhall2021nerf} leverages MLPs to model scenes volumetrically, achieving unprecedented rendering quality. However, these methods incur significant costs: slow training, high computational demands, and limited editability.

3D Gaussian Splatting \cite{kerbl20233d} overcomes these limitations through its explicit, Gaussian-based representation. By employing differentiable rasterization of 3D Gaussians, it achieves real-time rendering while maintaining NeRF-comparable visual quality. The discrete nature of Gaussians offers advantages in scene manipulation and surface reconstruction compared to continuous implicit representations. However, standard 3DGS prioritizes rendering quality over geometric precision, often producing misaligned Gaussian distributions that poorly approximate physical surfaces. Recent extensions like 2DGS \cite{huang20242d}, PGSR \cite{chen2024pgsr}, and GauSurf \cite{wang2024gaussurf} address this through geometric regularization, optimizing Gaussians to produce accurate depth and normal maps. While these methods improve surface estimation, they focus on aggregate outputs rather than individual Gaussian placement. Consequently, the Gaussians remain unevenly distributed and misaligned with underlying surfaces, limiting their effectiveness for geometry-sensitive applications including scene editing and dynamic object deformation.

Inspired by the foundational work on Point Set Surfaces \cite{pss2001}, we propose Gaussian Set Surface Reconstruction (GSSR), a method that optimizes Gaussian distributions by: (1) anchoring centers to latent scene surfaces, (2) enforcing spatial uniformity, and (3) aligning dominant normals with surface geometry. GSSR's explicit geometric formulation enables direct compatibility with point-based manipulation techniques, opening new possibilities for precise scene editing, animation workflows, and dynamic scene processing -- effectively bridging the gap between neural rendering quality and practical 3D content creation needs.

GSSR optimize individual Gaussian location and orientation through two key mechanisms: (1) The application of explicit geometric constraints derived from estimated planar structures and surface smoothness, encouraging Gaussians to coalesce onto latent scene surfaces; and (2) A joint optimization of opacity and position coupled with adaptive pruning, which aggressively eliminates Gaussians with low contributions while strategically repositioning others at underrepresented areas. Our experiments across multiple datasets demonstrate that our approach yields a cleaner scene representation with more geometrically consistent ellipsoids while preserving high rendering quality.

In summary, our key contributions include:
\begin{itemize}
\item GSSR, a 3DGS framework optimizing Gaussians for uniform placement and precise surface alignment
\item A geometric regularization technique that advances beyond prior 3DGS methods in per-Gaussian depth/normal accuracy and visual coherence
\item An opacity-position optimization strategy that prunes redundancies while enforcing surface adherence
\item State-of-the-art Gaussian accuracy and completeness across three datasets, without  rendering quality loss
\end{itemize}


\section{Related Work}
\subsection{Gaussian Splatting for 3D Reconstruction}
Due to the explicit Gaussian ellipsoids representation of 3DGS, many effective works\cite{guedon2024sugar,huang20242d,turkulainen2025dn,yu2024gsdf,chen2024pgsr,fan2024trim,chen2024vcr,wang2024gaussurf} that focus on applying geometric regularization to improve geometry accuracy are proposed. SuGaR~\cite{guedon2024sugar} introduces a signed distance function and density supervision to guide Gaussians toward object surfaces. 2DGS~\cite{huang20242d} replaces volumetric ellipsoids with the elliptical disk-shaped Gaussians, which better align with local geometry and improve surface fidelity during rendering. GOF~\cite{yu2024gaussian} integrates implicit neural representation within the Gaussian framework to enhance geometric detail and multi-view consistency. PGSR~\cite{chen2024pgsr} introduces a depth estimation strategy and geometric constraints that yield smoother and more accurate surface reconstruction. GausSurf~\cite{wang2024gaussurf} incorporates geometric guidance by combining patch-match MVS for texture-rich regions and normal priors for texture-less areas. Additionally, methods like GSDF~\cite{yu2024gsdf} and NeuSG~\cite{chen2023neusg} combine 3DGS with Signed Distance Field (SDF) networks to improve reconstruction quality. While these approaches effectively improve surface smoothness and depth accuracy, they often result in overly dense and irregular Gaussian distributions. In contrast, our method not only enhances surface reconstruction but also produces a cleaner and more spatially consistent Gaussian representation, while maintaining high-fidelity rendering quality.

\subsection{Scene Editing}
3D Gaussian Splatting offers photorealistic rendering and holds great promise for applications in XR, content creation, and digital twins. However, its ellipsoidal representation lacks geometric consistency with real-world surfaces, making intuitive and semantically meaningful scene editing difficult compared to mesh-based approaches. Recent methods such as GaussianEditor~\cite{chen2024gaussianeditor}, Gaussian Grouping~\cite{gs_grouping}, Point'n Move~\cite{point_n_move}, and Feng et al.~\cite{feng2024newsplitalgorithm3d} explore 3DGS editing through semantic prompts, mask-based grouping, or improved Gaussian splitting, but are largely limited to basic object-level operations such as removal, rotation, and translation. In parallel, some efforts have explored scene editing in NeRF-based representations~\cite{wang2023seal, huang2024seald, yuan2022nerf, haque2023instruct}, addressing similar challenges of geometry-awareness and user interaction. While promising, these approaches still face limitations due to implicit nature of NeRF and the lack of controllable structure. In this work, we aim to align the generated 3D Gaussians as closely as possible with the physical surfaces, thereby enabling more reliable and flexible editing using existing point-based manipulation techniques.

\section{3D Gaussian Splatting Preliminaries}
3D Gaussian Splatting builds on Elliptical Weighted Average (EWA) splatting~\cite{zwicker2001ewa}, extended with a differentiable formulation~\cite{kerbl20233d} to optimize both the number and parameters of Gaussians for scene representation. Each Gaussian is defined by its center $\bm{x} \in \mathbb{R}^3$, opacity $\alpha \in [0,1]$, covariance matrix in world space $\Sigma_i \in \mathbb{R}^{3 \times3}$, and view-dependent color via 16 SH coefficients. During rendering, the final pixel color $C$ is computed by compositing $N$ depth-sorted 2D Gaussians:
\begin{equation}
C(\bm{p}) = \sum_{i=1}^N T_i \alpha_i \bm{c}_i, \label{eq:C1-1} \quad T_i = \prod_{j=1}^{i-1} (1 - \alpha_j \mathcal{G}^{2D}_{j}(\bm{p})) 
\end{equation}
where $\bm{c}_i$ represents the view-dependent color, and $\mathcal{G}^{2D}_{j}$ is the projected 2D Gaussian distribution. To render depth, previous methods~\cite{cheng2024gaussianpro, jiang2024gaussianshader} replace $\bm{c}_i$ with the Gaussian center depth $z_i$ in Equation~\ref{eq:C1-1}. However, this yields biased, curved surfaces. Following PGSR \cite{chen2024pgsr}, we instead computed unbiased depth using Gaussian surface normals $\bm{n}_i$, aligned with the minimum scale axis. The per-Gaussian tangent plane distance is:
\begin{equation}
d_i = \left(R_c^\top (\bm{\mu}_i - \bm{t}_c)\right)^\top R_c^\top \bm{n}_i,
\end{equation}
where $R_c$ is the world-to-camera rotation and $\bm{t}_c$ is the camera origin. The global depth is then computed via ray-plane intersection:
\begin{equation}
D(p) = \frac{d(\bm{p})}{\bm{N}(\bm{p})^\top K^{-1} \tilde{\bm{p}}},
\end{equation}
with $K$ the intrinsic matrix and $\tilde{\bm{p}}$ the homogeneous pixel coordinate. This formulation ensures geometry-consistent depth estimation independent of Gaussian density.

\section{Methodology}
Given posed input images, we reconstruct a Gaussian Set Surface (GSS) – a collection of Gaussians that are uniformly distributed and precisely aligned with underlying scene surfaces. This representation combines the geometric manipulability of Point Set Surfaces~\cite{pss2001} with 3DGS's strengths in photorealistic novel view synthesis. Our Gaussian Set Surface reconstruction pipeline consists of three key components: (1) geometric regularization for accurate depth and normal estimation, (2) per-Gaussian optimization to refine individual geometric properties, and (3) enhanced density control enforcing uniform spatial distribution.

\subsection{Geometric Regularization}
\subsubsection{Flattening 3D Gaussian}
Accurately capturing real-world scene geometry using 3D Gaussians is inherently challenging. To better align with the true surface structure, the Gaussians are instead flattened into 2D representations, allowing them to more precisely conform to the underlying geometry of the scene. Rather than initializing with 2D Gaussian Splatting (2DGS) \cite{huang20242d}, a scale loss $L_s$ is introduced to progressively compress each Gaussian ellipsoid along its smallest scale axis. This process effectively flattens the ellipsoid into a plane that best approximates the underlying surface geometry. Following the approach in \cite{chen2023neusg, chen2024pgsr}, a penalty is applied directly to the minimum scale of each Gaussian to enforce flattening:
\begin{equation}
    \mathcal{L}_{\text{s}}  = \left\| \min(s_1, s_2, s_3) \right\|_1,
\end{equation}
where $s_1,s_2,s_3$ represents the scale parameters of each Gaussian ellipsoid.

\subsubsection{Single-View Normal Consistency}
\begin{figure}
    \centering
    \includegraphics[width=1\linewidth]{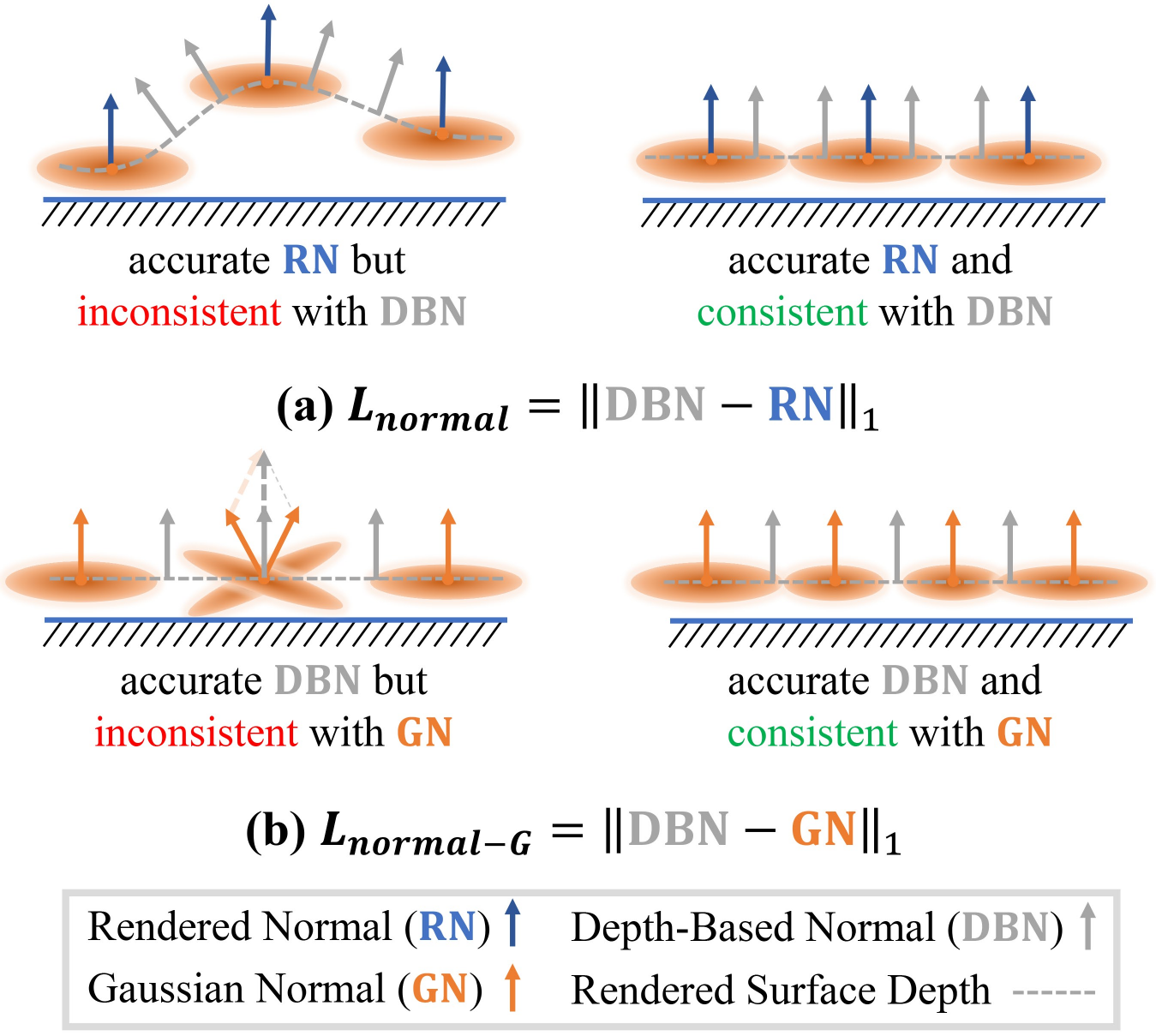}
    \caption{Illustration of (a) $L_{normal}$ and (b) $L_{normal-G}$. \textbf{Gaussian Normal} is derived from the direction of the Gaussian's minimum scale axis. \textbf{Rendered Normal} is computed via alpha blending of Gaussian normals along each pixel ray. \textbf{Depth-Based Normal} is estimated from the depth gradients of neighboring pixels. }   
    \label{fig:geometry_consistency}
\end{figure}

As illustrated in Figure~\ref{fig:geometry_consistency}(a), a locally discontinuous Gaussian-rendered plane may exhibit a smooth normal field, leading to inconsistency between local surface normals and depth geometry. To address this issue, encouraged by prior works~\cite{chen2024vcr, turkulainen2025dn, jiang2024gaussianshader, gao2024relightable}, a single-view normal loss is introduced to enforce local geometric consistency for every pixel $\bm{p}$ in the image domain $\Omega$:
\begin{equation}
\mathcal{L}_{\text{normal}} = \frac{1}{\Omega} \sum_{\bm{p} \in \Omega} \left\| \bm{n}_{\text{depth}}(\bm{p}) - \bm{n}_{\text{rendered}}(\bm{p}) \right\|_1,
\end{equation}
where $\bm{n}_{\text{depth}}(\bm{p})$ denotes the surface normal estimated from depth gradients of neighboring pixels, and $\bm{n}_{\text{rendered}}(\bm{p})$ denotes the normal rendered from the Gaussian sets.

\subsubsection{Multi-View Photometric Consistency}
\begin{figure}
    \centering
    \includegraphics[width=1\linewidth]{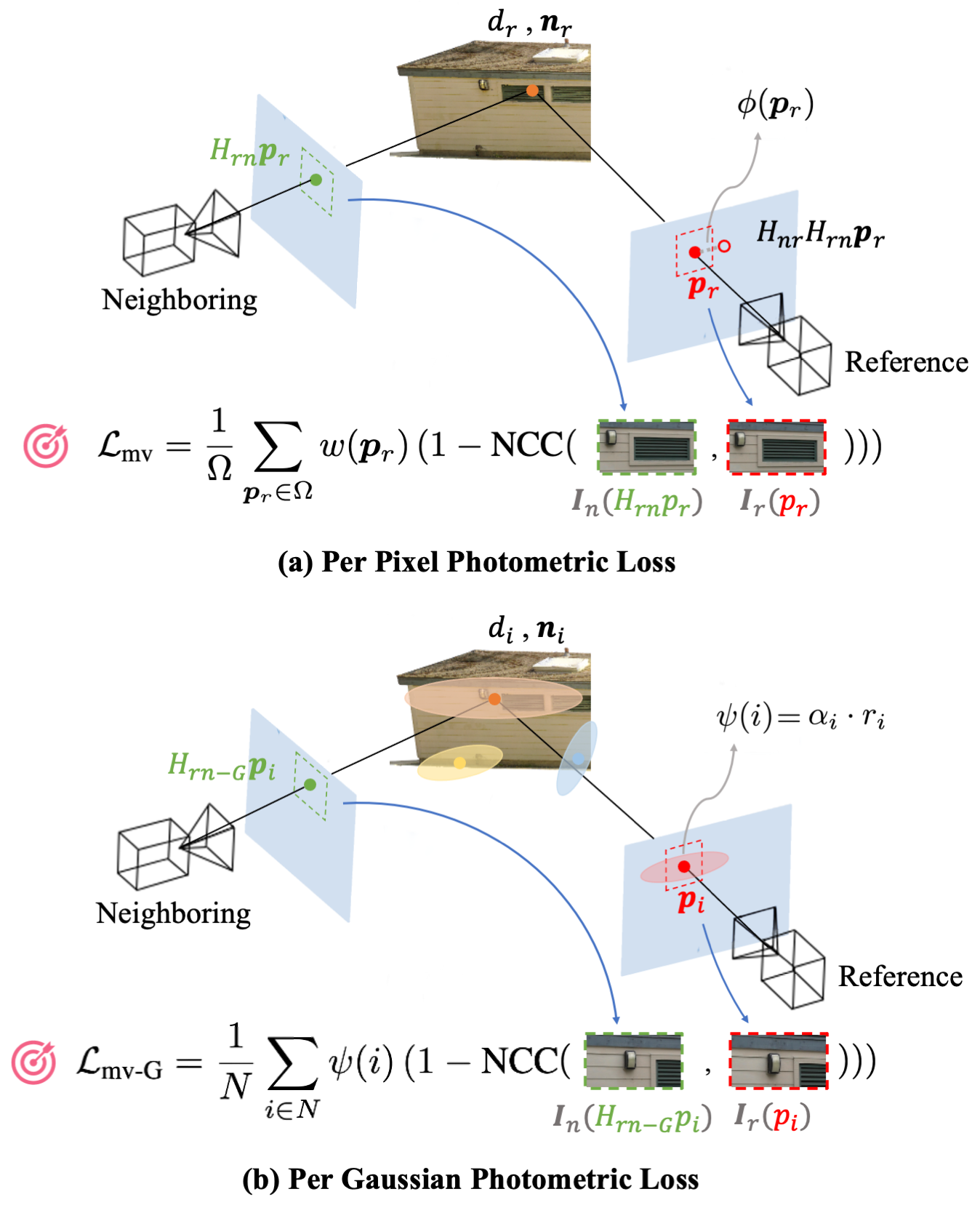}
    \caption{Illustration of pixel-level~(a) and Gaussian-level~(b) multi-view photometric loss. }
    \label{fig:mvloss}
\end{figure}

Photometric consistency is a widely used supervision signal in Multi-View Stereo (MVS) frameworks, leveraging color similarity across views to constrain geometry~\cite{galliani2016gipuma, fu2022geo, huang2025visibility}. Among various formulations, the Normalized Cross-Correlation (NCC) metric is particularly robust to illumination variation and exposure differences. In this work, the photometric NCC loss is adpoted to enforce multi-view consistency by comparing image patches in neighboring views. For each pixel $\bm{p}_r$ in the reference view, the corresponding pixel $\bm{p}_n$ in a neighboring view is computed using a plane-induced homography:
\begin{eqnarray}
    \bm{p}_n &=& H_{rn} \cdot \bm{p}_r,\label{C2-1} \\
    H_{rn} &=& K \left( R_{rn} - \frac{\bm{T}_{rn} \bm{n}_r^\top}{d_r} \right) K_r^{-1}, \label{C2-2}
\end{eqnarray}
where $R_{rn}$ and $\bm{T}_{rn}$ denote the relative rotation and translation from the reference view to the neighboring view, and $\bm{n}_r$, $d_r$ are the rendered surface normal and plane distance, respectively. As illustrated in Figure~\ref{fig:mvloss}(a), after computing the corresponding patch in the neighboring view, we followed the forward and backward projection error weighted NCC loss in PGSR~\cite{chen2024pgsr}:
\begin{align}
\phi(\bm{p}_r) &= \left\| \bm{p}_r - H_{nr} H_{rn} \bm{p}_r \right\| \label{eq:phi} \\[1ex]
w(p_r) &= 
\begin{cases}
\frac{1}{\exp(\phi(\bm{p}_r))}, & \text{if } \phi(\bm{p}_r) < 1 \\
0, & \text{if } \phi(\bm{p}_r) \geq 1
\end{cases} \label{eq:weight} \\[1ex]
\mathcal{L}_{\text{mv}} &= \frac{1}{\Omega} \sum_{\bm{p}_r \in \Omega} w(\bm{p}_r) \left( 1 - \text{NCC}(\bm{I}_r(\bm{p}_r), \bm{I}_n(H_{rn} \bm{p}_r)) \right) \label{eq:mvrgb},
\end{align}
where $\phi(\bm{p_r})$ represents the forward and backward projection error. If this error exceeds a predefined threshold, the pixel is considered occluded or associated with significant geometric inconsistency.

\subsection{Instance-Level Gaussian Optimization}
\begin{figure}
    \centering
    \includegraphics[width=1\linewidth]{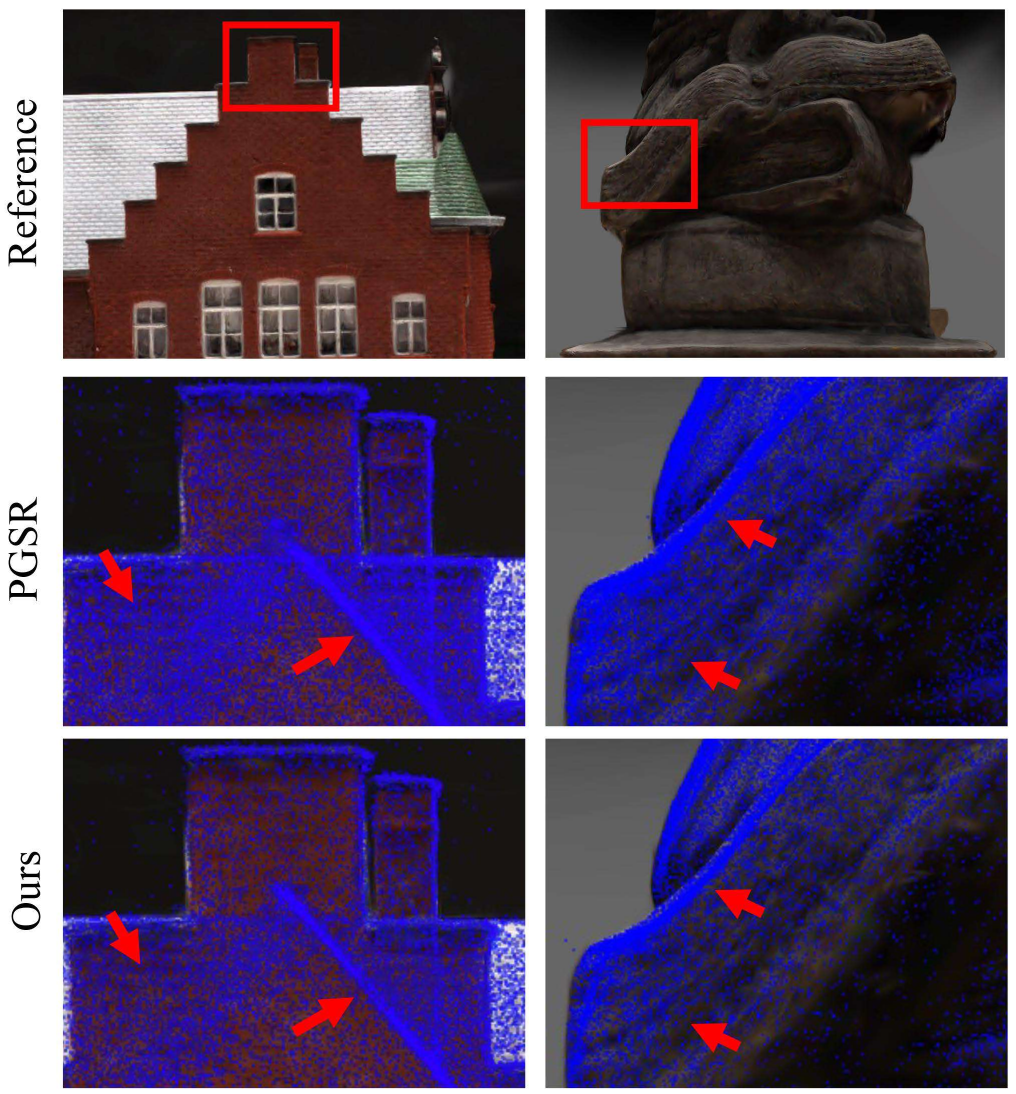}
    \caption{Comparison of Gaussian density distribution between PGSR and our method. Our approach produces lower Gaussian density, especially on planar regions (see red arrows), where the splats are more compact and narrowly distributed, demonstrating better geometric compactness and representation efficiency. }
    \label{fig:gaussian_density}
\end{figure}

Although geometric regularization enforces scene-level accuracy, it often fails to achieve satisfactory instance-level accuracy. Specifically, the Gaussians tend to form a thick layer surrounding the actual object surface, as illustrated in Figure~\ref{fig:gaussian_density}. This is primarily because the pixel-level geometric loss only constrains the alpha-blended depth and normal, without explicitly supervising individual Gaussian ellipsoids. To address this issue, we further incorporate geometric constraints to enforce single-view normal consistency and multi-view photometric consistency on each Gaussian instance. To filter out occluded Gaussians, we select only those whose center depth lies in front of the rendered depth map within a predefined tolerance for each training view.

\subsubsection{Bilaterally Weighted Normal Loss}
To enforce the single-view normal loss, a penalty is applied to the difference between each filtered Gaussian's normal and its depth-inferred normal at the center pixel. However, directly adding this loss can introduce instability during optimization, especially in the presence of incorrect, near-transparent, or very small Gaussians. To mitigate this issue, we adopt a bilaterally weighted normal loss, where the weights are determined by each Gaussian's opacity and splat size:

\begin{eqnarray}
\mathcal{L}_{\text{normal-G}} &=& \frac{1}{N} \sum_{i=1}^{N}  \psi(i) \left\| \bm{n}_{\text{depth}}(\bm{p}_i) - \bm{n}_i \right\|_1,\\
\psi(i) &=& \alpha_i \cdot r_i, 
\end{eqnarray}
where $\bm{p}_i$ denotes the 2D pixel location obtained by projecting the 3D center of the $i^\text{th}$ Gaussian onto the image plane, $\bm{n}_i$ is the Gaussian's normal in camera space, and $\psi(i)$ is a visibility-based weight defined as the product of opacity $\alpha_i$ and the projected splat radius $r_i$.

\subsubsection{Bilaterally Weighted Photometric Loss}
As illustrated in Figure~\ref{fig:mvloss}(b), the photometric NCC loss is employed to enforce multi-view consistency by comparing local image patches sampled around the projected centers of depth-filtered Gaussians. The same bilateral weight $\psi$ is applied to each Gaussian instance filtered by depth.
\begin{align}
 H_{rn-G} &= K \left( R_{rn} - \frac{\bm{T}_{rn} \bm{n}_i^\top}{d_i} \right) K_r^{-1}, \\[1ex]
\mathcal{L}_{\text{mv-G}} &= \frac{1}{N} \sum_{i \in N} \psi(i) \left( 1 - \text{NCC}(\bm{I}_r(\bm{p}_i), \bm{I}_n(H_{rn-G} \cdot \bm{p}_i)) \right) \label{eq:mvrgb},
\end{align}

This loss encourages the reconstructed geometry to align with phtometrically consistent regions across views, thereby improving the geometric accuracy of individual Gaussian instances.

\begin{table*}[!htbp]
    \centering
    \caption{Top: Reconstruction on DTU (Chamfer Distance $\downarrow$). Middle: Reconstruction on DTU (Accuracy of Gaussian Centroid $\downarrow$). Bottom: Reconstruction on DTU (Completeness of Gaussian $\downarrow$). \hlred{Red}, \hlorange{orange} and \hlyellow{yellow} backgrounds denote the \hlred{best}, \hlorange{second-best}, and \hlyellow{third-best} results respectively. }
    \label{tab:comp_recons_dtu}
    \begin{threeparttable}
    \resizebox{0.9\textwidth}{!}{
    \begin{tabular}{c|cccccccccccccccccc}
    \toprule
         & 24 & 37 & 40 & 55 & 63 & 65 & 69 & 83 & 97 & 105 & 106 & 110 & 114 & 118 & 122 & Mean \\
    \midrule
        
         2DGS\textsuperscript{*}~\cite{huang20242d} & 0.49 & \cellcolor{yellow!35}0.79 & \cellcolor{red!35}0.34 & 0.42 & \cellcolor{yellow!35}0.95 & 0.95 & 0.83 & 1.25 & 1.24 & \cellcolor{yellow!35}0.64 & 0.62 & 1.34 & \cellcolor{yellow!35}0.44 & 0.69 & 0.48 & \cellcolor{yellow!35}0.76 \\
         
         
         GOF\textsuperscript{*}~\cite{yu2024gaussian} & 0.49 & 0.83 & \cellcolor{orange!35}0.36 & 0.38 & 1.33 & \cellcolor{yellow!35}0.87 & \cellcolor{yellow!35}0.73 & 1.24 & 1.32 & 0.66 & 0.73 & 1.26 & 0.52 & 0.82 & 0.51 & 0.80 \\

    
         PGSR\textsuperscript{*}~\cite{chen2024pgsr} &  \cellcolor{orange!35}0.34 & \cellcolor{red!35}0.55 & 0.39 & \cellcolor{yellow!35}0.35 & \cellcolor{orange!35}0.78 & \cellcolor{red!35}0.58 & \cellcolor{red!35}0.49 & \cellcolor{red!35}1.09 & \cellcolor{red!35}0.63 & \cellcolor{red!35}0.59 & \cellcolor{orange!35}0.47 & \cellcolor{orange!35}0.50 & \cellcolor{red!35}0.30 & \cellcolor{red!35}0.37 & \cellcolor{red!35}0.34 & \cellcolor{red!35}0.52 \\

         GausSurf\textsuperscript{$\dagger$}~\cite{wang2024gaussurf} & \cellcolor{yellow!35}0.35 & \cellcolor{red!35}0.55 & \cellcolor{red!35}0.34 & \cellcolor{orange!35}0.34 & \cellcolor{red!35}0.77 & \cellcolor{red!35}0.58 & \cellcolor{orange!35}0.51 & \cellcolor{orange!35}1.10 & \cellcolor{yellow!35}0.69 & \cellcolor{orange!35}0.60 & \cellcolor{red!35}0.43 & \cellcolor{red!35}0.49 & \cellcolor{orange!35}0.32 & \cellcolor{yellow!35}0.40 & \cellcolor{yellow!35}0.37 & \cellcolor{red!35}0.52 \\
         
         Ours & \cellcolor{red!35} 0.33 & \cellcolor{orange!35}0.57 & \cellcolor{yellow!35}0.37 & \cellcolor{red!35}0.33 & \cellcolor{orange!35}0.78 & \cellcolor{orange!35}0.62 & \cellcolor{orange!35}0.51 & \cellcolor{yellow!35}1.11 & \cellcolor{orange!35}0.68 & \cellcolor{red!35}0.59 & \cellcolor{yellow!35}0.48 & \cellcolor{yellow!35}0.55 & \cellcolor{red!35}0.30 & \cellcolor{orange!35}0.38 & \cellcolor{orange!35}0.35 & \cellcolor{orange!35}0.53 \\

    \midrule
         2DGS\textsuperscript{*} & \cellcolor{yellow!35}0.66 & \cellcolor{yellow!35}1.02 & \cellcolor{orange!35}0.55 & \cellcolor{orange!35}0.60 & \cellcolor{yellow!35}0.88 & \cellcolor{yellow!35}1.03 & \cellcolor{yellow!35}0.92 & \cellcolor{yellow!35}0.55 & \cellcolor{yellow!35}0.95 & \cellcolor{yellow!35}0.47 & \cellcolor{yellow!35}0.52 & \cellcolor{yellow!35}0.99 & \cellcolor{yellow!35}0.63 & \cellcolor{yellow!35}0.42 & \cellcolor{yellow!35}0.41 & \cellcolor{yellow!35}0.71 \\
         GOF\textsuperscript{*} & 0.98 & 1.28 & 0.99 & 0.85 & 1.11 & 1.29 & 1.08 & 0.77 & 1.10 & 0.62 & 0.65 & 1.05 & 0.79 & 0.48 & 0.61 & 0.91 \\
         PGSR\textsuperscript{*} & \cellcolor{orange!35}0.58 & \cellcolor{orange!35}0.68 & \cellcolor{yellow!35}0.85 & \cellcolor{yellow!35}0.71 & \cellcolor{orange!35}0.74 & \cellcolor{orange!35}0.64 & \cellcolor{orange!35}0.68 & \cellcolor{orange!35}0.51 & \cellcolor{orange!35}0.74 & \cellcolor{orange!35}0.45 & \cellcolor{orange!35}0.46 & \cellcolor{orange!35}0.60 & \cellcolor{orange!35}0.46 & \cellcolor{orange!35}0.37 & \cellcolor{orange!35}0.37 & \cellcolor{orange!35}0.59 \\
         Ours & \cellcolor{red!35}0.32 & \cellcolor{red!35}0.47 & \cellcolor{red!35}0.46 & \cellcolor{red!35}0.29 & \cellcolor{red!35}0.56 & \cellcolor{red!35}0.42 & \cellcolor{red!35}0.39 & \cellcolor{red!35}0.43 & \cellcolor{red!35}0.52 & \cellcolor{red!35}0.33 & \cellcolor{red!35}0.26 & \cellcolor{red!35}0.33 & \cellcolor{red!35}0.21 & \cellcolor{red!35}0.26 & \cellcolor{red!35}0.25 & \cellcolor{red!35}0.37 \\
    
    \midrule
         2DGS\textsuperscript{*} & 0.92 & \cellcolor{yellow!35}0.87 & 1.02 & 0.72 & 1.06 & 1.47 & 1.25 & 1.94 & \cellcolor{yellow!35}1.70 & 1.11 & 1.56 & 1.51 & 0.81 & 1.33 & 1.09 & 1.22 \\
         GOF\textsuperscript{*} & \cellcolor{orange!35}0.71 & \cellcolor{orange!35}0.72 & \cellcolor{orange!35}0.81 & \cellcolor{orange!35}0.59 & \cellcolor{yellow!35}0.96 & \cellcolor{yellow!35}1.33 & \cellcolor{yellow!35}1.12 & \cellcolor{yellow!35}1.78 & \cellcolor{yellow!35}1.70 & \cellcolor{yellow!35}0.99 & \cellcolor{yellow!35}1.38 & \cellcolor{yellow!35}1.14 & \cellcolor{yellow!35}0.69 & \cellcolor{yellow!35}1.19 & \cellcolor{yellow!35}0.92 & \cellcolor{yellow!35}1.07 \\
         PGSR\textsuperscript{*} & \cellcolor{yellow!35}0.75 & \cellcolor{orange!35}0.72 & \cellcolor{yellow!35}0.88 & \cellcolor{yellow!35}0.64 & \cellcolor{orange!35}0.82 & \cellcolor{orange!35}1.16 & \cellcolor{orange!35}0.97 & \cellcolor{orange!35}1.69 & \cellcolor{orange!35}1.44 & \cellcolor{orange!35}0.96 & \cellcolor{orange!35}1.30 & \cellcolor{orange!35}0.90 & \cellcolor{orange!35}0.68 & \cellcolor{orange!35}1.05 & \cellcolor{orange!35}0.86 & \cellcolor{orange!35}0.99 \\
         Ours & \cellcolor{red!35}0.62 & \cellcolor{red!35}0.61 & \cellcolor{red!35}0.58 & \cellcolor{red!35}0.51 & \cellcolor{red!35}0.56 & \cellcolor{red!35}0.88 & \cellcolor{red!35}0.67 & \cellcolor{red!35}1.52 & \cellcolor{red!35}1.17 & \cellcolor{red!35}0.82 & \cellcolor{red!35}0.85 & \cellcolor{red!35}0.54 & \cellcolor{red!35}0.46 & \cellcolor{red!35}0.64 & \cellcolor{red!35}0.55 & \cellcolor{red!35}0.73 \\
         
    \bottomrule
    \end{tabular}}

    \begin{tablenotes}[para]
    \item[*] \textit{ Reproduced results using the authors' official implementation.}

    \item[$\dagger$] \textit{The source code is not available; only mesh-based evaluation results are reported.}
    \end{tablenotes}

    \end{threeparttable}
    
\end{table*}


\begin{table}[!htbp]
    \centering
    \caption{Top: Reconstruction on TnT (F1 Score $\uparrow$). Middle: Reconstruction on TnT (Precision of Gaussian Centroid $\uparrow$). Bottom: Reconstruction on TnT (Completeness of Gaussian $\uparrow$).}
    \label{tab:comp_recons_tnt}
    \begin{threeparttable}
    \resizebox{0.48\textwidth}{!}{
    \begin{tabular}{c|ccccccc}
    \toprule
         & Barn & Caterpillar & Courthouse & Ignatius & Meetingroom & Truck & Mean \\
    \midrule
         2DGS\textsuperscript{*}~\cite{huang20242d} & 0.46 & 0.24 & 0.15 & 0.49 & 0.19 & 0.45 & 0.33  \\
         
         GOF\textsuperscript{*}~\cite{yu2024gaussian} & \cellcolor{yellow!35}0.55 & \cellcolor{yellow!35}0.39 & \cellcolor{orange!35}0.28 & 0.71 & 0.26 & 0.57 & \cellcolor{yellow!35}0.46  \\
         
         PGSR\textsuperscript{*}~\cite{chen2024pgsr} & \cellcolor{red!35}0.65 & \cellcolor{red!35}0.45 & 0.21 & \cellcolor{red!35}0.81 & \cellcolor{orange!35}0.33 & \cellcolor{orange!35}0.62 & \cellcolor{red!35}0.51  \\
         
         GausSurf\textsuperscript{$\dagger$}~\cite{wang2024gaussurf} & 0.50 & \cellcolor{orange!35}0.42 & \cellcolor{red!35}0.30 & \cellcolor{yellow!35}0.73 & \cellcolor{red!35}0.39 & \cellcolor{red!35}0.65 & \cellcolor{orange!35}0.50 \\
         
         Ours & \cellcolor{orange!35}0.64 & \cellcolor{orange!35}0.42 & \cellcolor{yellow!35}0.22 & \cellcolor{orange!35}0.80 & \cellcolor{yellow!35}0.32 & \cellcolor{yellow!35}0.59 & \cellcolor{orange!35}0.50 \\
    \midrule
         2DGS\textsuperscript{*} & \cellcolor{orange!35}0.65 & \cellcolor{yellow!35}0.59 & \cellcolor{red!35}0.62 & \cellcolor{yellow!35}0.71 & \cellcolor{orange!35}0.47 & \cellcolor{orange!35}0.71 & \cellcolor{yellow!35}0.62 \\
         GOF\textsuperscript{*}  & \cellcolor{orange!35}0.65 & \cellcolor{orange!35}0.60 & \cellcolor{red!35}0.62 & \cellcolor{orange!35}0.77 & \cellcolor{red!35}0.49 & \cellcolor{yellow!35}0.70 & \cellcolor{orange!35}0.64  \\
         PGSR\textsuperscript{*} & \cellcolor{yellow!35}0.63 & 0.58 & \cellcolor{orange!35}0.56 & 0.70 & \cellcolor{red!35}0.49 & 0.69 & 0.61 \\
         Ours & \cellcolor{red!35}0.66 & \cellcolor{red!35}0.66 & \cellcolor{yellow!35}0.52 & \cellcolor{red!35}0.83 & \cellcolor{orange!35}0.47 & \cellcolor{red!35}0.78 & \cellcolor{red!35}0.65 \\   
    \midrule
         2DGS\textsuperscript{*} & 0.08 & 0.07 & \cellcolor{yellow!35}0.04 & 0.09 & 0.04 & 0.11 & 0.07 \\
         GOF\textsuperscript{*}  & \cellcolor{yellow!35}0.12 & \cellcolor{yellow!35}0.09 & \cellcolor{orange!35}0.05 & \cellcolor{yellow!35}0.15 & \cellcolor{yellow!35}0.05 & \cellcolor{yellow!35}0.16 & \cellcolor{yellow!35}0.10  \\
         PGSR\textsuperscript{*} & \cellcolor{orange!35}0.16 & \cellcolor{orange!35}0.13 & \cellcolor{red!35}0.08 & \cellcolor{orange!35}0.16 & \cellcolor{red!35}0.09 & \cellcolor{orange!35}0.20 & \cellcolor{orange!35}0.14 \\
         Ours & \cellcolor{red!35}0.20 & \cellcolor{red!35}0.19 & \cellcolor{red!35}0.08 & \cellcolor{red!35}0.28 & \cellcolor{orange!35}0.07 & \cellcolor{red!35}0.32 & \cellcolor{red!35}0.19 \\
    \bottomrule
    \end{tabular}}


    \end{threeparttable}

\end{table}

\begin{figure*}
    \centering
    \includegraphics[width=0.9\textwidth]{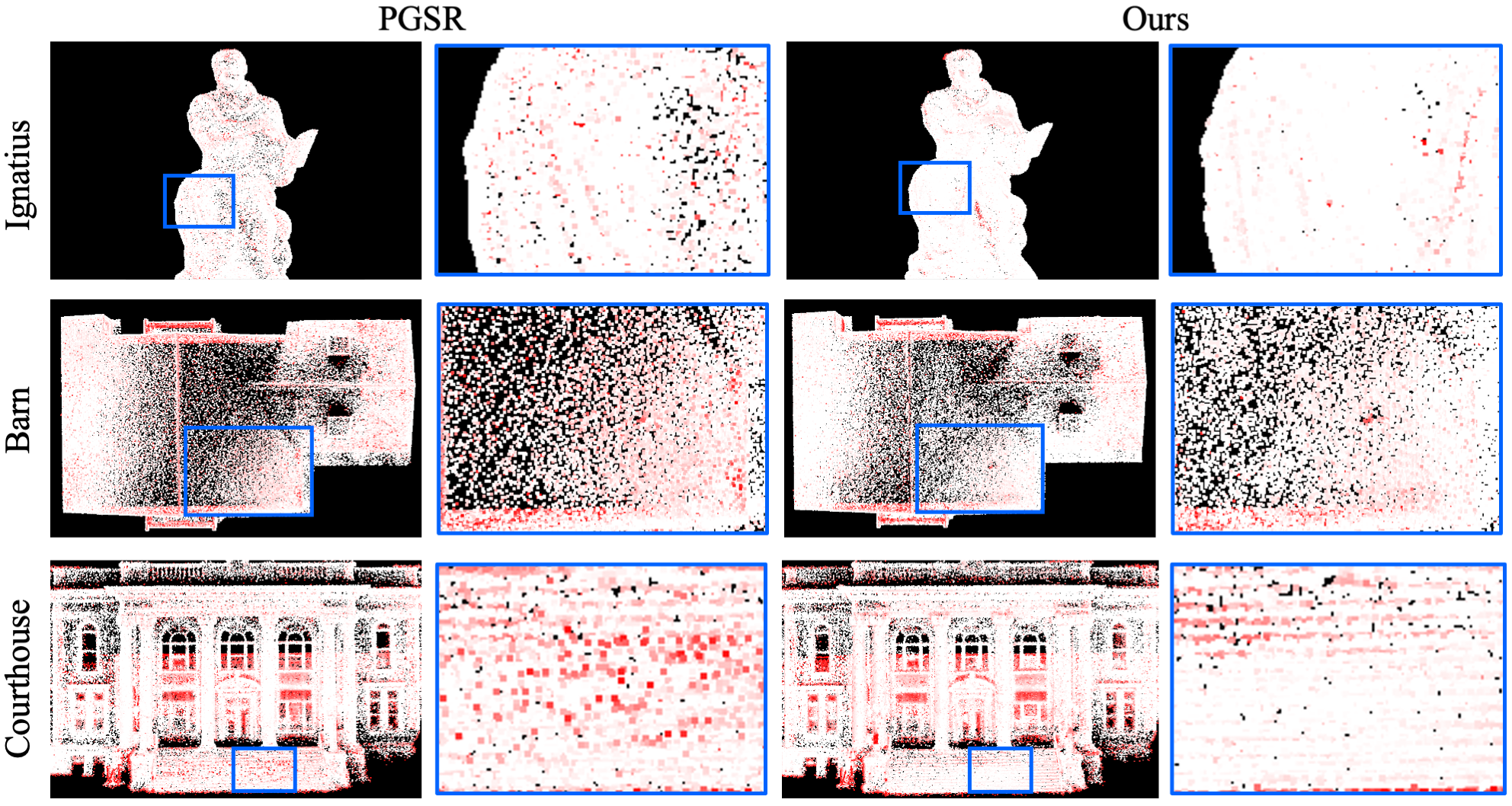}
    \caption{Comparison of Gaussian centroid precision on TnT. \textbf{Less} \textcolor{red}{red} points in reconstructed scenes indicate \textbf{higher accuracy}.}
    \label{fig:tnt_comparison}
\end{figure*}



\subsection{Gaussian Density Control}
In this framework, the original gradient-based Gaussian densification scheme from 3DGS is retained, while an additional opacity loss and Gaussian resampling strategy are introduced to perform density control. 

\subsubsection{Depth Filtered Opacity Loss}
One major reason that Gaussians form a thick layer surrounding the actual object surface is that many of them remain semi-transparent. To address this issue, we introduce an opacity regularization loss that encourages each Gaussian's opacity to converge toward either 0 (fully transparent and removable) or 1 (fully opaque):
\begin{equation}
\mathcal{L}_{\text{opacity}} = \frac{1}{N} \sum_{i=1}^{N} \left( \log \alpha_i + \log (1 - \alpha_i) \right),
\end{equation}

\subsubsection{Depth \& Normal Reinitialization}

To adaptively control the spatial distribution of Gaussians, we draw inspiration from the periodic resampling strategy in Mini-Splatting~\cite{fang2024mini}, extending it with a view-based, opacity-guided approach. Unlike Mini-Splatting, which does not explicitly consider visibility or density imbalance, our method leverages per-view opacity statistics to prevent oversampling in view-crowded regions and ensures that new Gaussians are allocated more evenly across the scene. For each training view, we compute the accumulated opacity $\alpha_{\text{acc}}$ by rendering all visible Gaussians, \ie, those positioned in front of the rendered depth map. Regions with $\alpha_{\text{acc}} \gg 1$ indicate areas that are unnecessarily oversampled.  The sampling weight at each pixel $p$ is therefore defined as the inverse transmittance: $1-\alpha_{\text{acc}}(p)$. The number of newly sampled Gaussians for the view is determined by:
\begin{equation}
N_{\text{new}} = N_{\text{per-view}} \cdot \frac{1}{|\Omega|} \sum_{p \in \Omega} \left(1 - \alpha_{\text{acc}}(p)\right),
\end{equation}
where $N_{\text{per-view}}$ is a predefined constant that controls the base number of Gaussians sampled per view. To prioritize under-represented regions, 3D points are drawn from the rendered point cloud via multinomial sampling, with probabilities proportional to the sampling weights. For each selected point, its 3D position and normal are extracted from the rendered depth and normal maps, and then transformed to world coordinates. To enhance robustness, a bilateral filter is applied to the depth and normal maps, incorporating both spatial proximity and value similarity. New Gaussians are initialized at these sampled points are flattened according to the corresponding normals. 


\section{Experiments}
This section presents an evaluation of the proposed method, including implementation details and ablation studies. \textbf{The source code will be released upon the acceptance of the paper.}

\subsection{Datasets and Implementation}
\textbf{Datasets:}
We evaluate our method on three datasets: DTU~\cite{jensen2014large}, Tanks and Temples (TnT)~\cite{knapitsch2017tanks} and Mip-NeRF360~\cite{barron2022mip_nerf360}, covering both indoor and outdoor environments. The DTU dataset consists of 124 sets of high-quality images of various objects with complex geometries and textures, captured under controlled lighting with camera poses. The Tanks and Temples dataset serves as a benchmark for complex large-scale scene reconstruction using high-resolution video input. Following PGSR~\cite{chen2024pgsr}, we evaluate 3D reconstruction performance on 15 DTU scenes and 6 Tanks and Temples scenes. To assess novel view synthesis, we use the Mip-NeRF360 dataset, which contains high-resolution images of unbounded outdoor scenes.


\begin{table*}[!htbp]
    \centering
    \caption{Quantitative results on the Mip-NeRF360~\cite{barron2022mip_nerf360} dataset.}
    \label{tab:comp_rendering}
    \begin{threeparttable}
    \resizebox{0.8\textwidth}{!}{
    \begin{tabular}{c|ccc|ccc|ccc}
    \toprule
        \multirow{2}*{Method} & \multicolumn{3}{c}{Indoor scenes} & \multicolumn{3}{c}{Outdoor scenes} & \multicolumn{3}{c}{Average on all scenes} \\
        & PSNR $\uparrow$   & SSIM $\uparrow$   & LPIPS $\downarrow$ & PSNR $\uparrow$ & SSIM $\uparrow$ & LPIPS $\downarrow$ & PSNR $\uparrow$ & SSIM $\uparrow$ & LPIPS $\downarrow$ \\
    \midrule
        2DGS~\cite{huang20242d} & \cellcolor{orange!35}30.40 & 0.916 & 0.195 & 24.34 & 0.717 & 0.246 & 27.37 & 0.817 & 0.221\\
        GOF~\cite{yu2024gaussian} & \cellcolor{red!35}30.79 & \cellcolor{orange!35}0.924 & \cellcolor{yellow!35}0.184 & \cellcolor{orange!35}24.82 & \cellcolor{yellow!35}0.750 & \cellcolor{red!35}0.202 & \cellcolor{red!35}27.81 & \cellcolor{orange!35}0.837 & \cellcolor{orange!35}0.193 \\
        PGSR\textsuperscript{*}~\cite{chen2024pgsr} & \cellcolor{yellow!35}30.20 & \cellcolor{red!35}0.930 & \cellcolor{red!35}0.158 & \cellcolor{yellow!35}24.77 & \cellcolor{orange!35}0.752 & \cellcolor{orange!35}0.204 & \cellcolor{yellow!35}27.49 & \cellcolor{red!35}0.841 & \cellcolor{red!35}0.181 \\
        GausSurf~\cite{wang2024gaussurf} & 30.05 & \cellcolor{yellow!35}0.920 & 0.183 & \cellcolor{red!35}25.09 & \cellcolor{red!35}0.753 & \cellcolor{yellow!35}0.212 & \cellcolor{orange!35}27.57 & \cellcolor{orange!35}0.837 & \cellcolor{yellow!35}0.198 \\
        Ours & 29.79 & \cellcolor{orange!35}0.924 & \cellcolor{orange!35}0.168 & 24.47 & 0.724 & 0.250 & 27.13 & \cellcolor{yellow!35}0.824 & 0.209 \\
    \bottomrule
    \end{tabular}}
    \end{threeparttable}
\end{table*}

\textbf{Implementation:}
The implementation of our method is based on PyTorch. All experiments are conducted on a desktop equipped with an NVIDIA RTX 4090 GPU. Our training strategy and hyperparameters are generally consistent with previous works 3DGS~\cite{kerbl20233d} and PGSR~\cite{chen2024pgsr}. All scenes are trained for 30,000 iterations. The final training loss is defined as:
\begin{align}
\mathcal{L} &= \mathcal{L}_{\text{RGB}} 
+ \lambda_1 \mathcal{L}_{\text{normal}} 
+ \lambda_2 \mathcal{L}_{\text{normal-G}} \nonumber \\
&\quad + \lambda_3 \mathcal{L}_{\text{opacity}} 
+ \lambda_4 \mathcal{L}_{\text{mv}} 
+ \lambda_5 \mathcal{L}_{\text{mv-G}}.
\end{align}

We set the weights as follows: $\lambda_1=0.015$, $\lambda_2=0.0075$, $\lambda_3=0.0001$, $\lambda_4=0.15$, and $\lambda_5=0.15$. Depth and normal reinitialization are performed at iterations 5,000 and 10,000. We set the number of Gaussian sampled per-view $N_{\text{per-view}}=10,000.$

\subsection{Results}
\textbf{Reconstruction:}
To assess overall reconstruction accuracy, we use the F1 score on the Tanks and Temples dataset and the Chamfer Distance on the DTU dataset. We compare the reconstruction performance of GSSR against recent GS-based methods, including 2DGS~\cite{huang20242d}, GOF~\cite{yu2024gaussian}, PGSR~\cite{chen2024pgsr}, GausSurf~\cite{wang2024gaussurf}. For evaluation, we first render a depth map from each training view and then apply Truncated Signed Distance Function (TSDF) Fusion~\cite{newcombe2011kinectfusion} to integrate them into a unified TSDF field. A mesh is subsequently extracted using the Marching Cube algorithm~\cite{lorensen1998marching}, which is used to compute the reconstruction metrics. As shown in the top rows of Tables~\ref{tab:comp_recons_dtu} and~\ref{tab:comp_recons_tnt}, our method achieves performance comparable to state-of-the-art results on both datasets.

\textbf{Gaussian Instance Accuracy:}
To assess the geometric fidelity of individual Gaussians, we report accuracy (average point-to-ground-truth distance) and completeness (average ground-truth-to-point distance) as the two components of Chamfer Distance on the DTU dataset. Similarly, we report precision and recall as components of the F1 score for the Tanks and Temples dataset. As shown in Table~\ref{tab:comp_recons_dtu} and Table~\ref{tab:comp_recons_tnt}, GSSR consistently outperforms all other methods in Gaussian-level geometric quality. Figure~\ref{fig:teaser} provides a visual comparison on the DTU dataset between GSSR, 2DGS~\cite{huang20242d}, and PGSR~\cite{chen2024pgsr}, where GSSR yields both lower position error and more complete coverage. Figure~\ref{fig:tnt_comparison} shows Gaussian center precision across three Tanks and Temples scenes, demonstrating GSSR's higher accuracy. Figure~\ref{fig:gaussian_density} visualizes the spatial density of Gaussians, where GSSR produces a more uniform distribution and thinner layers aligned with object surfaces.

\begin{figure*}[t!]
    \centering
    \includegraphics[width=\linewidth]{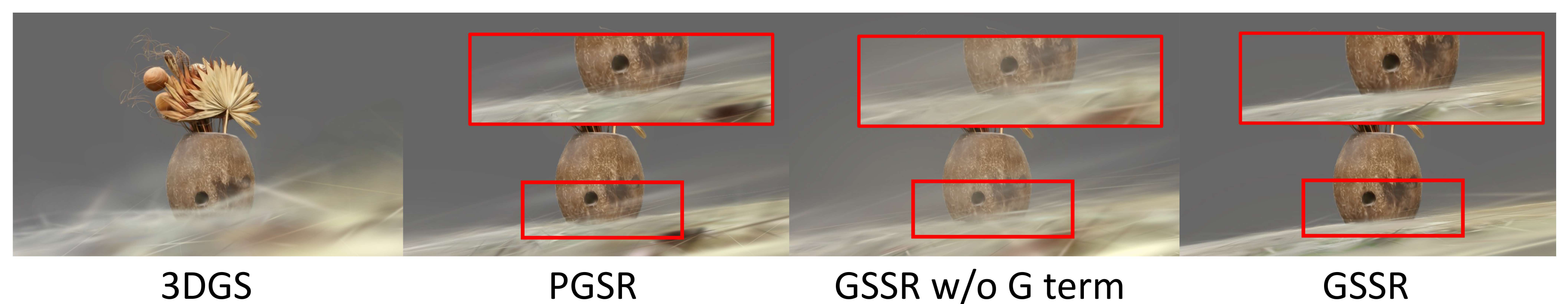}
    \caption{Scene editing comparison: placing an object on a flat surface. GSSR yields the cleanest composition.}
    \label{fig:main_editing}
\end{figure*}

\textbf{Novel View Synthesis:} 
To evaluate novel view synthesis quality, we follow the experimental setup of 3DGS~\cite{kerbl20233d} and conduct validation on the Mip-NeRF360 dataset~\cite{barron2022mip_nerf360}. Our method is compared against several GS-based approaches. As shown in Table~\ref{tab:comp_rendering}, the proposed framework not only achieves a strong surface reconstruction performance, but also delivers competitive results in novel view rendering.

\textbf{Scene Editing}
By design, GSSR produces Gaussians that are both aligned with and evenly distributed along the latent surfaces. 
As showsn in Figure~\ref{fig:main_editing}, when placing an additional object on a surface, our method yields cleaner and more coherent compositions, where PGSR~\cite{chen2024pgsr} and our ablation variants introduce noticeable artifacts. More examples are provided in supplementary material.

\begin{table}
    \centering
    \caption{Ablation study on the DTU dataset.}
    \resizebox{0.47\textwidth}{!}{
    \begin{tabular}{|c|ccc|}
    \hline
        Method &  Chamfer Distance $\downarrow$ & Precision $\downarrow$ & Completeness $\downarrow$\\
    \hline
        w/o Resampling & 0.57 & 0.48 & 0.92 \\
        w/o $L_{normal}$ &  \cellcolor{orange!35}0.53  & 0.40 & \cellcolor{red!35}0.68 \\
        w/o $L_{mv}$ & 0.63 & 0.43 & 0.85 \\
        w/o $L_{s}$ & \cellcolor{yellow!35}0.54 &\cellcolor{orange!35} 0.38 & 0.76 \\
        w/o $L_{mv-G}$, $L_{normal-G}$ & \cellcolor{red!35}0.51 & 0.56 & \cellcolor{yellow!35}0.75 \\
        Full model & \cellcolor{orange!35}0.53 & \cellcolor{red!35}0.37 & \cellcolor{orange!35}0.73 \\
    \hline
    \end{tabular}}
    \label{tab:Ablation}
\end{table}

\subsection{Ablation Study}
To further evaluate the effectiveness of key components in our proposed method, we perform ablation studies on the DTU dataset. We report the quality of the reconstructed mesh, as well as the accuracy and completeness of the Gaussian centers. Detailed quantitative results for each ablated variant are presented in Table~\ref{tab:Ablation}. The resampling strategy plays a critical role in improving completeness by reducing under-coverage. Loss terms $\mathcal{L}_{\text{normal-G}}$, and $\mathcal{L}_{\text{mv-G}}$ significantly enhance Gaussian instance accuracy, while the multi-view consistency loss $\mathcal{L}_{\text{mv}}$ is crucial for improving overall geometric quality. Overall, the full model consistently achieves the best trade-off across all metrics, demonstrating the importance of each component in enhancing geometric fidelity and scene coverage.


\section{Conclusion}
We present Gaussian Set Surface Reconstruction (GSSR), a novel approach inspired by Point Set Surfaces~\cite{pss2001} that represents scenes using dense, geometrically-precise 3D Gaussians with uniform spatial distribution. GSSR enforces geometric accuracy through multi-scale constraints (pixel-level and Gaussian-level) while preserving photorealistic rendering capabilities. Our framework introduces: (1) an opacity regularization loss to prune redundant Gaussians, and (2) a view-adaptive resampling strategy for optimal spatial distribution. Comprehensive evaluation across three real-world datasets demonstrates GSSR's superior geometric consistency and rendering fidelity compared to state-of-the-art 3DGS methods, producing cleaner distributions that better adhere to scene geometry.

\paragraph{Limitation:}
While GSSR significantly improves per-Gaussian accuracy, two limitations remain: First, these gains do not directly translate to improved mesh reconstruction or alpha-blended depth quality. Second, the view-based Gaussian sampling requires empirical parameter setting, though this provides coarse control over Gaussian density. Future work will investigate adaptive sampling strategies to automate this process.


{\small
\bibliographystyle{ieee_fullname}
\bibliography{egbib}
}

\end{document}


\title{Supplementary Material }  

\maketitle
\thispagestyle{empty}
\appendix


\section{Primitive Analysis \& Scene Editing Examples}


With improved primitive-level accuracy, our Gaussians adhere closely to the underlying object surface, rather than drifting into surrounding space. Figure~\ref{fig:thin_plane} highlights this difference on a Tanks and Temples scene~\cite{knapitsch2017tanks}: while our method produces primitives that remain directly anchored to the true geometry, PGSR~\cite{chen2024pgsr} forms a thick shell of Gaussians with inconsistent normal around the surface. This is especially evident on Figure~\ref{fig:thin_plane}(d) where PGSR forms a layer band of artifacts. In these regions, crossed Gaussians with misaligned normals partially counterweigh each other, which explains why the fused depth and normals may still appear plausible despite primitive-level inaccuracies. 

Such misalignments not only inflate the representation but also lead to noisy and unstable local geometry. By enforcing Gaussian-level constraints during optimization, GSSR yields a tighter and more surface-consistent primitive distribution. The advantages extend to downstream applications such as scene editing. For example in Figure~\ref{fig:editing} we place additional object on the surface. Because our Gaussians form a cleaner and more accurately aligned surface, the composition is sharper and more coherent, whereas competing methods introduce surface irregularities that reduce realism.


\section{Additional Experiments}

\subsection{Training Time and Memory Usage}

Table~\ref{tab:efficiency} compares the efficiency of different Gaussian-based representations in terms of the number of Gaussians, storage size, and training time. Our method maintains competitive efficiency while providing stronger geometric accuracy. Specifically, although the number of Gaussians used in our approach is higher than 2DGS~\cite{huang20242d} and GOF~\cite{yu2024gaussian}, it remains comparable to PGSR~\cite{chen2024pgsr}, and the associated memory and training cost are within the same range. Importantly, our design achieves this balance without relying on a hard cap on the number of primitives, instead leveraging depth- and normal-guided reintialization to keep the representation compact and surface-consistent. This shows that our improvements in accuracy and editability come at only a modest increase in resource usage.

\begin{table}[h]
    \centering
    \begin{tabular}{lccc}
            & \# Primitives & Size & Time\\
            \hline
        PGSR~\cite{chen2024pgsr} & 1120k & 277MB & 38 mins\\
        GOF~\cite{yu2024gaussian} & 869K & 219MB & 61 mins\\
        2DGS~\cite{huang20242d} & 504k & 124MB & 15 mins\\
        Ours & 995k & 247MB & 39 mins\\
    \end{tabular}
    \caption{Efficiency comparison on an example scene from the Tanks and Temples dataset~\cite{knapitsch2017tanks}. Our method achieves improved primitive accuracy while keeping the number of Gaussians, memory usage, and training time comparable to prior work.}
    \label{tab:efficiency}
\end{table}

\subsection{Number of Gaussians}

\begin{figure}[h]
    \centering
    \includegraphics[width=\linewidth]{rebuttal_img/GS_num.png}
    \caption{Sensitivity study of $N_{\text{per-view}}$ on the DTU dataset. Increasing $N_{\text{per-view}}$ reduces Chamfer Distance (better geometry) but increases the number of Gaussians.}
    \label{fig:num}
\end{figure}

Thanks to our depth \& normal reinitialization, we can regulate the number of Gaussians without imposing a strict upper bound. To better assess the sensitivity of the parameter  $N_{\text{per-view}}$, which controls the base number of Gaussians initialized per view, we conduct a study on how this parameter affects both geometric accuracy and the final number of Gaussians. Figure~\ref{fig:num} presents the results: as $N_{\text{per-view}}$ increases, the Chamfer Distance decreases at the beginning, indicating improved geometric accuracy, whilc the total number of Gaussians grows steadily. Beyond a certain point (around 10k), the accuracy curve begins to saturate, whereas the number of Gaussians continues to rise. This demonstrates that our method can flexibly balance accuracy and efficiency: small $N_{\text{per-view}}$ value yield a compact representation with moderate accuracy, while larger values provide tighter surface alignment at the cost of more primitives. Importantly, our reinitialization avoids uncontrolled oversampling, ensuring that even with higher $N_{\text{per-view}}$ the growth of Gaussians remains stable and meaningful.

\begin{figure*}[t!]
    \centering
    \includegraphics[width=\linewidth]{rebuttal_img/ThinPlane.pdf}
    \caption{Comparison of primitive placement: GSSR aligns Gaussians tightly with the surface, while PGSR and GSSR without Gaussian-level constraints forms a thick shell with inconsistent normals. In (a) and (b), which show different viewing angles of the rooftop, our approach produce a smoother and more coherent layer of Gaussians, while competing methods exhibit jagged edges. A similar effect is visible on the ground in (c) and (d), where GSSR maintains a cleaner, surface-aligned distribution. }
    \label{fig:thin_plane}
\end{figure*}

\subsection{Depth \& Normal Map Comparison}

Figure~\ref{fig:normal_depth} presents depth and normal visualizations across several Tanks and Temples scenes. At the fused map level, GSSR achieves same level of results with PGSR~\cite{chen2024pgsr}, and our ablation variant (GSSR without Gaussian-level optimization). This demonstrates that our method does not compromise multi-view consistency. However, it is important to note that fused maps alone cannot reveal local irregularities in Gaussian placement or normal alignment. For instance, PGSR may produce a smooth fused mesh in under-constrained areas, even though the underlying primitives form thick layers with crossed normals. Our primitive-level evaluations (see previous section) provide complementary evidence that GSSR achieves a more surface-consistent distribution of Gaussians, which is not directly visible in fused depth and normal maps.

\subsection{Accumulated Opacity After Reinitialization}

To further examine the effect of our reinitialization strategy, we visualize the accumulated opacity of Gaussians after reinitialization. As shown in Figure~\ref{fig:Alpha_map}, which presents examples from multiple scenes of the DTU dataset, the accumulated maps demonstrate that our primitives are evenly distributed across the object surfaces. This leads to balanced opacity accumulation, ensuring that the Gaussian representation remains compact while maintaining consistent surface coverage.


\begin{figure*}
    \centering
    \includegraphics[width=\linewidth]{rebuttal_img/BetterSceneEditing.pdf}
    \caption{Scene editing example: placing an object on the surface produces clean composition with GSSR.}
    \label{fig:editing}
\end{figure*}

\begin{figure*}[h]
    \centering
    \includegraphics[width=\linewidth]{rebuttal_img/BetterNormalDepth.pdf}
    \caption{Depth and normal map comparisons on several Tanks and Temples scenes. Our methods produce competitive results at the fused depth/normal level. However, these maps do not fully expose local normal inconsistencies or per-Gaussian misalignments.}
    \label{fig:normal_depth}
\end{figure*}

\begin{figure*}[h]
    \centering
    \includegraphics[width=\linewidth]{rebuttal_img/Alpha_map.png}
    \caption{Visualization of the accumulated opacity after reinitialization. }
    \label{fig:Alpha_map}
\end{figure*}



{\small
\bibliographystyle{ieee_fullname}
\bibliography{egbib}
}